\documentclass[conference]{IEEEtran}
\IEEEoverridecommandlockouts

\usepackage{cite}
\usepackage{microtype}
\usepackage{amsmath,amssymb,amsfonts}
\usepackage{mathtools}
\usepackage{amsthm}
\usepackage{algorithm}
\usepackage{algorithmic}
\usepackage{graphicx}
\usepackage{subcaption}
\usepackage{booktabs}
\usepackage{multirow}
\usepackage{colortbl}
\usepackage{threeparttable}
\usepackage{textcomp}
\usepackage[hidelinks]{hyperref}
\usepackage{placeins}

\def\BibTeX{{\rm B\kern-.05em{\sc i\kern-.025em b}\kern-.08em
    T\kern-.1667em\lower.7ex\hbox{E}\kern-.125emX}}

\theoremstyle{plain}

\theoremstyle{definition}

\theoremstyle{remark}

\begin{document}

\title{PeakFocus: Bridging Peak Localization and Intensity Regression via a Unified Multi-Scale Framework for Electricity Load Forecasting}

\author{
\IEEEauthorblockN{Wangzhi Yu\IEEEauthorrefmark{1}, Peng Zhu\IEEEauthorrefmark{1}, Qing Zhao\IEEEauthorrefmark{1}, Yiwen Jiang\IEEEauthorrefmark{2}, Dawei Cheng\IEEEauthorrefmark{1}}
\IEEEauthorblockA{
\IEEEauthorrefmark{1}\textit{School of Computer Science and Technology, Tongji University}, Shanghai, China\\
wangzhiy@tongji.edu.cn, pengzhu@tongji.edu.cn, 2452445@tongji.edu.cn, dcheng@tongji.edu.cn
}
\IEEEauthorblockA{\IEEEauthorrefmark{2}\textit{Big Data Center, State Grid Corporation of China}, Beijing, China\\
yiwen-jiang@sgcc.com.cn}
}

\maketitle

\begin{abstract}
 Electricity load peak forecasting (ELPF), simultaneously predicting peak timing and intensity, is a prerequisite for effective grid scheduling and risk management. However, existing methods face three limitations. First, they adopt a two-stage predict-then-locate paradigm, which severs the link between temporal localization and intensity regression. Second, they still struggle with the multi-scale representation conflict, leading to peak misjudgment and timing misalignment. Third, the lack of explicit peak timing context during intensity regression causes intensity smoothing because predictions are dominated by global smoothing trends. To address these limitations, we propose PeakFocus, a unified framework for ELPF. (i) A Unified Peak-Aware Pipeline (UPAP) utilizes a triple hybrid loss to jointly supervise temporal localization and intensity regression, alongside a tolerance-based evaluation protocol. (ii) A Multi-Scale Mixing Peak Locator (MSM-PL) exploits coarse-grained features to mitigate peak misjudgment caused by local fluctuations, and injects them into fine-grained features via a cascade mechanism to resolve timing misalignment. (iii) A Location-Aware Decoder (LAD) injects peak timing context into the intensity regression process, providing explicit guidance to counteract intensity smoothing and improve peak intensity estimation. Extensive experiments on the public Electricity (ELC) dataset and our industrial-scale World Large-scale Electricity Load (WLEL) dataset show that PeakFocus outperforms baselines in both timing precision and intensity estimation. Code and model implementation are available at \url{https://github.com/TongjiFinLab/PeakFocus}.
\end{abstract}


%%%%%%%%%%%%%%%%%%%%%%%%%%%%%%%% Introduction %%%%%%%%%%%%%%%%%%%%%%%%%%%%%%%%
\section{Introduction}
\label{sec:intro}

In industrial power grids, effective resource management relies heavily on forecasting electricity load demand. Within such load signals, peaks are rare structural events sparsely embedded within otherwise smooth temporal patterns, which makes their accurate forecasting fundamentally different from learning dominant trends. Within this scope, accurate peak forecasting is of central operational importance, as it must estimate both the timing and the intensity of the next load peak. Accurate timing informs grid operators when to bring reserves online and dispatch supplementary generation, while accurate intensity determines how much reserve capacity is required to safely accommodate the upcoming demand surge. Yet traditional deep time-series models such as Transformers~\cite{zhou2021informer, wu2021autoformer} and Multi-Layer Perceptrons (MLPs)~\cite{wang2024timemixer, hu2025adaptive} are trained with global reconstruction losses~\cite{wang2024deep} that minimize average error, biasing predictions toward overly smooth trends~\cite{benidis2022deep} that flatten extreme values and miss the very peaks that matter. Electricity Load Peak Forecasting (ELPF)~\cite{amara2023daily, huang2025metaeformer} has therefore emerged as a specialized task targeting critical peak events.

% ================= FIGURE INSERTION =================
\begin{figure}[t]
  \centering
  \includegraphics[width=\columnwidth]{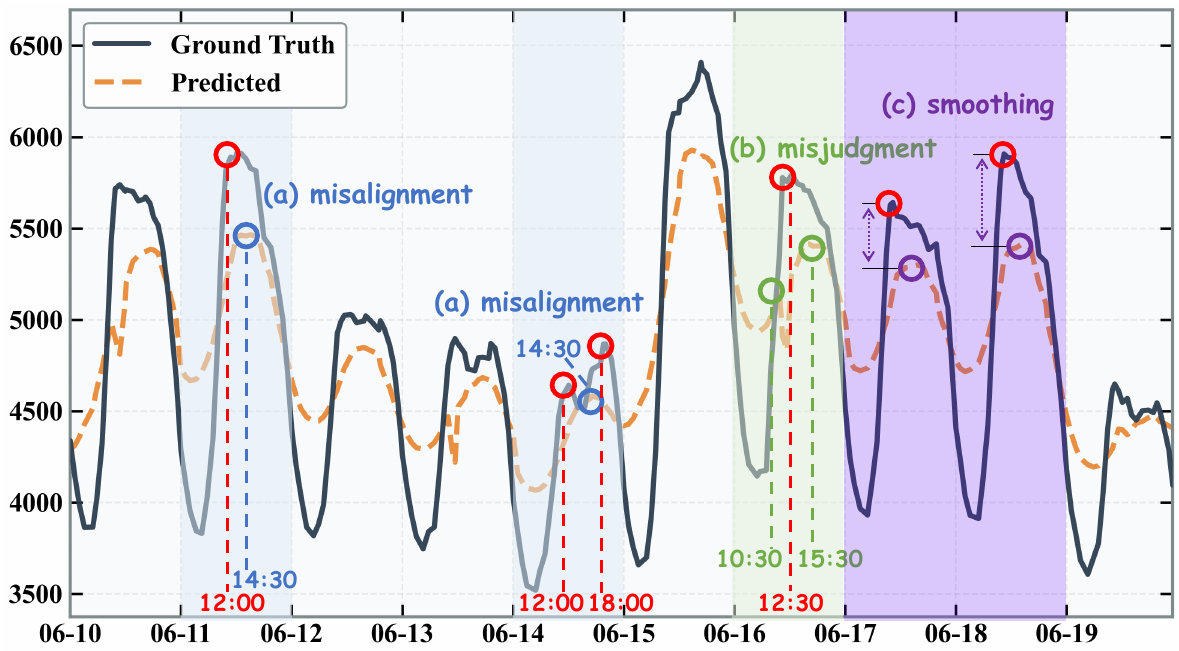}
  \vspace{-13pt}
  \caption{An illustration of limitations in PatchTST for ELPF under practical forecasting scenarios.
  (a)~Timing Misalignment and (b)~Peak Misjudgment show the multi-scale representation conflict, where coarse-grained features lack temporal resolution (causing shifts) while fine-grained features are easily affected by local fluctuations (leading to false positives).
  (c)~Intensity Smoothing shows the underestimation caused by the lack of explicit peak timing context, where the intensity decoder lacks this context and is pulled toward global smoothing trends.}
  \label{fig:intro_failures}
  \vspace{-20pt}
\end{figure}
% ================= FIGURE INSERTION END =================

Existing ELPF methods have clear structural limitations~\cite{dai2021electrical}.
First, when applying TSF models, they typically rely on a two-stage predict-then-locate paradigm.
This separation breaks the link between temporal localization and intensity regression, preventing the model from learning their relationship effectively.
Some end-to-end methods have been proposed~\cite{zhang2023unlocking, zhu2025enhancing} to bridge this gap. Second, they still struggle with the multi-scale representation conflict.
Coarse-grained features lack temporal resolution, resulting in timing misalignment (Figure~\ref{fig:intro_failures}a), while fine-grained features are affected by local fluctuations, leading to peak misjudgment (Figure~\ref{fig:intro_failures}b).
Third, intensity regression suffers from a lack of explicit peak timing context. Even in unified frameworks, the decoder operates without this signal and predictions are pulled toward global smoothing trends (Figure~\ref{fig:intro_failures}c), often leading to peak underestimation.
This makes reliable temporal localization and accurate intensity estimation particularly challenging in real-world grid-scale forecasting deployment.\looseness=-1

To address these limitations, we propose PeakFocus, a unified framework for ELPF that jointly models temporal localization and intensity regression. To bridge their separation, the Unified Peak-Aware Pipeline (UPAP) introduces a triple hybrid objective that couples temporal localization and intensity regression during training. To resolve the multi-scale representation conflict, the Multi-Scale Mixing Peak Locator (MSM-PL) leverages coarse-grained information to suppress local fluctuations and propagates it to fine-grained features through a top-down cascade for accurate temporal alignment. To reduce intensity smoothing, the Location-Aware Decoder (LAD) injects peak timing context into regression, enabling more accurate intensity estimation in practical grid settings without being dominated by global smoothing trends. Together, these components form a unified framework that jointly addresses the three structural bottlenecks identified above. This coupling not only sharpens timing precision but also tightens intensity estimation. Building on this coordinated design, our key contributions summarized as follows:
\vspace{-1pt}
\begin{itemize}
  \setlength{\itemsep}{1pt}
  \setlength{\parsep}{0pt}
  \setlength{\topsep}{-2pt}
    \item \textbf{Unified End-to-End Framework Bridging Localization and Regression.} We propose a fundamental paradigm shift from conventional global reconstruction to sparse extreme value optimization. By bridging temporal localization and numerical regression, our framework unifies peak timing localization and intensity regression. This methodology enables the direct learning of critical peak signals typically suppressed by standard global objectives.
    
    \item \textbf{Multi-Scale Peak-Aware Architectural Modules.} We design specialized components to tackle intrinsic structural conflicts. Specifically, the MSM-PL resolves the trade-off between misjudgment and misalignment via multi-scale mixing, while the LAD employs context injection to overcome the intensity smoothing dilemma.
    \item \textbf{Tolerance-Aware Evaluation Protocol with an Industrial-Grid Benchmark.} We establish a unified, tolerance-aware evaluation protocol to jointly assess both localization accuracy and intensity regression error metrics. PeakFocus is validated on the public \textsc{ELC} benchmark and our industrial-scale \textsc{WLEL} dataset (2021--2025, real regional grid), with consistent gains across both.
\end{itemize}

%%%%%%%%%%%%%%%%%%%%%%%%%%%%%%%% Preliminaries and Problem Definition %%%%%%%%%%%%%%%%%%%%%%%%%%%%%%%%
\section{Preliminaries and Problem Definition}
\label{sec:preliminaries}

\subsection{Peak Definition}
\label{sec:peak_def}
We characterize peak events based on local optimality and structural prominence. Given the ground-truth load intensity series $\mathbf{Y}_{i\_\mathrm{true}} \in \mathbb{R}^{H \times 1}$ over a forecasting horizon of $H$ time steps, the corresponding peak indicator series $\mathbf{Y}_{t\_\mathrm{true}} \in \{0,1\}^{H \times 1}$ is generated offline by the standard peak detector from the open-source Python library \texttt{findpeaks}\footnote{\url{https://github.com/erdogant/findpeaks}}. Formally, a time step $t$ is labeled as a peak ($\mathbf{Y}_{t\_\mathrm{true}}[t]=1$) only when two conditions are jointly satisfied under a sensitivity threshold $\eta$ and a lookahead window $\ell$: (i) the signal at $t$ is a local maximum from which the subsequent values drop by at least $\eta$, and (ii) no larger value appears within the following $\ell$ steps. A symmetric rise of at least $\eta$ above a local minimum marks the transition back from valley tracking. This hysteresis-style rule suppresses spurious local fluctuations while preserving structurally prominent peaks.

\subsection{ELPF Task Formulation}
\label{sec:task_formulation}
The goal of ELPF is to jointly anticipate the timing and intensity of peak events. Formally, given a historical univariate load series $\mathbf{X}_{1:T} \in \mathbb{R}^{T \times 1}$ of length $T$ together with timestamp features $\mathbf{M}_{1:T+H}$ spanning both the historical and forecasting windows, the model learns a mapping $\mathcal{F}_{\theta}$ parameterized by $\theta$ that produces two aligned outputs over a forecasting horizon of $H$ time steps:
\begin{equation}
\label{eq:task_mapping}
[\mathbf{Y}_{i\_\mathrm{pred}}, \mathbf{Y}_{t\_\mathrm{pred}}] = \mathcal{F}_\theta(\mathbf{X}_{1:T}, \mathbf{M}_{1:T+H}),
\end{equation}
where $\mathbf{Y}_{i\_\mathrm{pred}} \in \mathbb{R}^{H \times 1}$ is the predicted intensity sequence and $\mathbf{Y}_{t\_\mathrm{pred}} \in [0,1]^{H \times 1}$ collects the per-step peak occurrence probabilities.

%%%%%%%%%%%%%%%%%%%%%%%%%%%%%%%% Methodology %%%%%%%%%%%%%%%%%%%%%%%%%%%%%%%%
\section{Methodology}
\label{sec:methodology}

\begin{figure*}[ht]
  \centering
  \includegraphics[width=\textwidth]{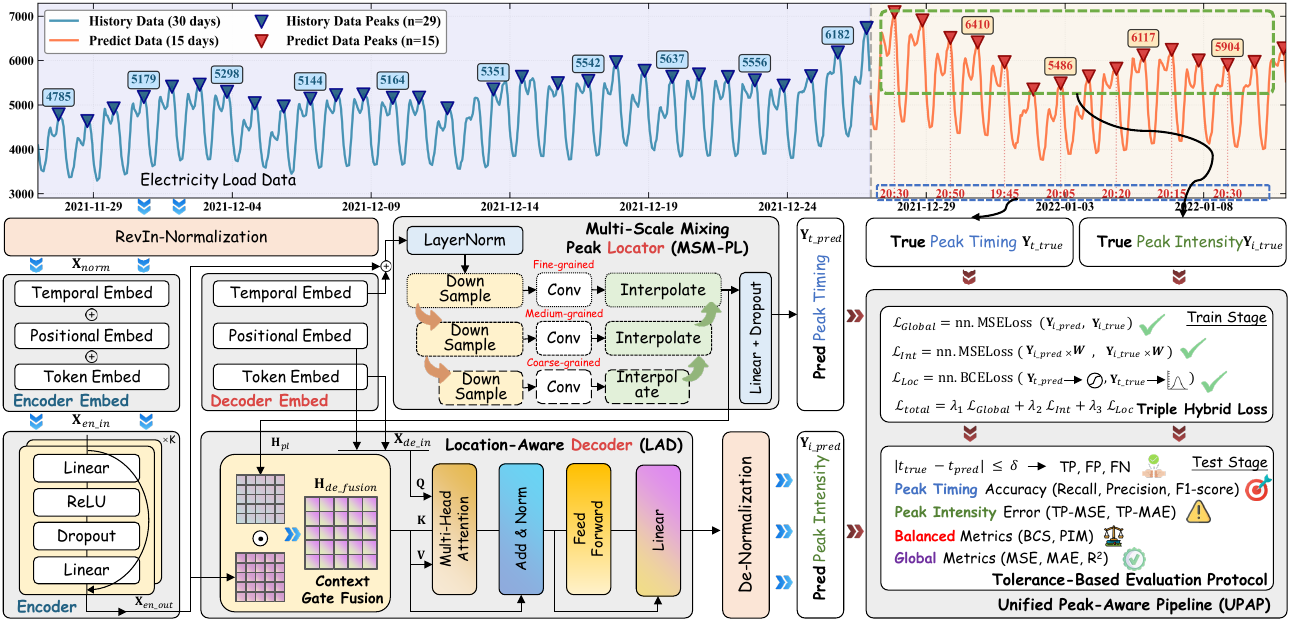}
  \vspace{-5pt}
  \caption{
    The proposed PeakFocus architecture.
    An encoder extracts input features.
    MSM-PL resolves localization conflicts via multi-scale mixing, outputting peak states $\mathbf{H}_{\mathrm{pl}}$ and timing $\mathbf{Y}_{t\_\mathrm{pred}}$.
    LAD prevents intensity smoothing by conditioning regression $\mathbf{Y}_{i\_\mathrm{pred}}$ on $\mathbf{H}_{\mathrm{pl}}$ via Context Gate Fusion.
    UPAP ensures robust optimization via a Triple Hybrid Objective ($\mathcal{L}_{\mathrm{total}}$) under tolerance-aware event-level semantics and soft peak supervision.
    }
  \label{fig:peakfocus}
  \vspace{-15pt}
\end{figure*}

As illustrated in Figure~\ref{fig:peakfocus}, PeakFocus is designed as a dual-head architecture underpinned by a shared backbone. 
The process begins with the Encoder (Section~\ref{sec:encoder}), which applies Reversible Instance Normalization (RevIN) followed by data embedding to raw inputs $\mathbf{X}_{1:T}$ and temporal features $\mathbf{M}_{1:T}$, constructing a horizon-aligned context.
To simultaneously capture both timing and intensity, the architecture branches into two specialized streams. 
First, the MSM-PL (Section~\ref{sec:msm_pl}) injects coarse-grained semantics into fine-grained features, producing the peak hidden state $\mathbf{H}_{\mathrm{pl}}$ and timing probabilities $\mathbf{Y}_{t\_\mathrm{pred}}$. 
Leveraging this context, the LAD (Section~\ref{sec:lad}) explicitly conditions the intensity regression $\mathbf{Y}_{i\_\mathrm{pred}}$ on $\mathbf{H}_{\mathrm{pl}}$ via location-aware cross-attention to prevent intensity smoothing. 
Finally, the entire framework is optimized end-to-end via the UPAP (Section~\ref{sec:upap}) using a triple hybrid objective under consistent tolerance-aware semantics.

%%%%%%%%%%%%%%%%%%%%%%%%%%%%%%%%%%%%%%%%%%%%%%%%%%%%%%%%%%%%%%%%
%%%%%%%%%%%%%%%%%%%%%%%%%%%%%%%%%%%%%%%%%%%%%%%%%%%%%%%%%%%%%%%%

%%%%%%%%%%%%%%%%%%%%%%%%%%%%%%%% Normalization and Encoder %%%%%%%%%%%%%%%%%%%%%%%%%%%%%%%%

\subsection{Normalization and Encoder}
\label{sec:encoder}

As illustrated in Figure~\ref{fig:peakfocus}, we first apply RevIN normalization~\cite{kim2021reversible} to the input $\mathbf{X}_{1:T}\in\mathbb{R}^{T\times 1}$ to obtain the corresponding zero-mean unit-variance normalized series:
\begin{equation}
\label{eq:revin}
\mathbf{X}_{\mathrm{norm},1:T}=\operatorname{RevIN}(\mathbf{X}_{1:T})
\end{equation}

The input data for the encoder $\mathbf{X}_{\mathrm{en\_in}}$ are generated by the Data Embedding Layer~\cite{zhou2021informer}, including value projection, positional encoding, and temporal features:
\begin{equation}
\label{eq:x-en-in}
\mathbf{X}_{\mathrm{en\_in}}
=
\mathcal{E}_{\mathrm{val}}(\mathbf{X}_{\mathrm{norm},1:T})
+
\mathcal{E}_{\mathrm{pos}}
+
\mathcal{E}_{\mathrm{time}}(\mathbf{M}_{1:T})
% \in\mathbb{R}^{T\times d}.
\end{equation}
where $\mathcal{E}_{\mathrm{val}}(\cdot)$ is the 1D convolutional value embedding, $\mathcal{E}_{\mathrm{pos}}$ the fixed positional encoding, and $\mathcal{E}_{\mathrm{time}}(\cdot)$ the timestamp embedding applied to $\mathbf{M}_{1:T}$.

A residual MLP backbone \( f_{\mathrm{enc}} \)~\cite{shao2022spatial} of latent dimension $d$, followed by a learnable projection \( \Pi \), then produces the horizon-aligned context \( \mathbf{X}_{\mathrm{en\_out}} \):
\begin{align}
\label{eq:enc}
\mathbf{Z}_{\mathrm{hist}}
&=
f_{\mathrm{enc}}(\mathbf{X}_{\mathrm{en\_in}})
\in\mathbb{R}^{T\times d} \\
\label{eq:temporal-proj}
\mathbf{X}_{\mathrm{en\_out}}
&=
\Pi(\mathbf{Z}_{\mathrm{hist}})
\in\mathbb{R}^{H\times d}
\end{align}

%%%%%%%%%%%%%%%%%%%%%%%%%%%%%%%% Multi-Scale Mixing Peak Locator %%%%%%%%%%%%%%%%%%%%%%%%%%%%%%%%

\subsection{Multi-Scale Mixing Peak Locator (MSM-PL)}
\label{sec:msm_pl}

MSM-PL is designed to resolve the intrinsic trade-off between peak misjudgment and timing misalignment.
We incorporate future temporal information into the peak branch by injecting time features into the shared context~\cite{hu2025adaptive}, followed by layer normalization~\cite{wu2024role} to stabilize the feature distribution.
Note that unlike the temporal embedding in the encoder (Eq.~\ref{eq:x-en-in}), which encodes historical time context $\mathbf{M}_{1:T}$, the injection here introduces future time features $\mathbf{M}_{T+1:T+H}$ to provide positional awareness of the forecasting horizon. This design enables the peak locator to distinguish periodic patterns, which is critical for accurate peak timing. The peak-branch input $\mathbf{H}_{\mathrm{in}}^{p} \in \mathbb{R}^{H \times d}$ is computed as:
\begin{equation}
\label{eq:future-inj}
\mathbf{H}_{\mathrm{in}}^{p}
=
\operatorname{LayerNorm}\left(
\mathbf{X}_{\mathrm{en\_out}}
+
\mathcal{E}_{\mathrm{time}}(\mathbf{M}_{T+1:T+H})
\right)
\end{equation}

To mitigate misjudgments caused by local fluctuations, we employ a hierarchical architecture comprising a bottom-up pyramid and a top-down cascade, drawing on the multi-scale feature fusion paradigm established in computer vision~\cite{lin2017feature, ronneberger2015unet}.
Inspired by~\cite{wu2021autoformer}, we construct a $K$-layer hierarchy of coarse-grained features through iterative downsampling, where $K$ denotes the pyramid depth. 
Let $\operatorname{AvgPool}_{\kappa, s}(\cdot)$ denote the average pooling operation parameterized by kernel size $\kappa$ and stride $s$ (we use $s$ to avoid conflict with the sigmoid function $\sigma(\cdot)$ used later). 
The pyramid generation is formally defined as the following recursion:
\begin{equation}
\label{eq:pyr}
\mathbf{S}^{(0)}=\mathbf{H}_{\mathrm{in}}^{p}\qquad
\mathbf{S}^{(k)}=\operatorname{AvgPool}_{\kappa^{(k)}, s^{(k)}}(\mathbf{S}^{(k-1)})
\end{equation}
To leverage these robust semantics for precise alignment, we fuse the coarse representations back into finer scales.
This is achieved via a cascade injection mechanism based on a convolutional layer $\psi_k(\cdot)$ (with kernel size $\kappa_c$) and an average interpolation upsampling operator $\mathcal{U}_k(\cdot)$.
The cascade generation is performed recursively for $k=K-1,\dots,0$ as:
\begin{align}
\label{eq:cascade-base}
\mathbf{F}^{(K)} &= \psi_K(\mathbf{S}^{(K)}) \\
\label{eq:cascade-step}
\mathbf{F}^{(k)} &= \psi_k(\mathbf{S}^{(k)}) + \mathcal{U}_k(\mathbf{F}^{(k+1)})
\end{align}
The cascade injection mechanism performs top-down hierarchical fusion: the coarsest scale $\mathbf{F}^{(K)}$ suppresses local fluctuations to capture global semantics; each subsequent finer scale integrates its local features with the upsampled coarser representation, providing structural transition and precise temporal alignment.
The resulting finest fused feature $\mathbf{F}^{(0)}$ constitutes the peak hidden state $\mathbf{H}_{\mathrm{pl}} \in \mathbb{R}^{H \times d}$, serving as context for the downstream decoder.
We obtain the peak timing probability $\mathbf{Y}_{t\_\mathrm{pred}}$ by projecting this representation through a linear classifier parameterized by weight $\mathbf{W}_{p}\in\mathbb{R}^{d\times 1}$ and bias $\mathbf{b}_{p}\in\mathbb{R}$, followed by the sigmoid activation function $\sigma(\cdot)$:
\begin{equation}
\label{eq:y-t-pred}
\mathbf{Y}_{t\_\mathrm{pred}}
=
\sigma(\mathbf{H}_{\mathrm{pl}}\mathbf{W}_{p}+\mathbf{b}_{p})
\in[0,1]^{H\times 1}
\end{equation}

%%%%%%%%%%%%%%%%%%%%%%%%%%%%%%%% Location-Aware Decoder %%%%%%%%%%%%%%%%%%%%%%%%%%%%%%%%

\subsection{Location-Aware Decoder (LAD)}
\label{sec:lad}

To counteract the intensity smoothing dilemma, the LAD explicitly conditions the regression process on the peak hidden state $\mathbf{H}_{\mathrm{pl}}$ via a location-aware cross-attention mechanism.
We first construct decoder queries $\mathbf{X}_{\mathrm{de\_in}}$ by embedding normalized values with positional encoding:
\begin{equation}
\label{eq:x-de-in}
\mathbf{X}_{\mathrm{de\_in}}
=
\mathcal{E}_{\mathrm{val}}(\mathbf{X}_{\mathrm{dec}})
+
\mathcal{E}_{\mathrm{pos}}
\in\mathbb{R}^{H\times d}
\end{equation}
where $\mathbf{X}_{\mathrm{dec}} \in \mathbb{R}^{H \times C}$ denotes the raw decoder input values, constructed by concatenating the last $L_{\mathrm{label}}$ values of the input series with a zero-padded placeholder for the forecasting horizon, following the standard encoder-decoder paradigm~\cite{zhou2021informer}. $\mathcal{E}_{\mathrm{pos}}$ provides the localized context.
Simultaneously, to inject peak timing context into the shared horizon representation, we employ a Context Gate Fusion mechanism to modulate $\mathbf{X}_{\mathrm{en\_out}}$ using the peak hidden state $\mathbf{H}_{\mathrm{pl}}$, where $\odot$ denotes element-wise multiplication:
\begin{equation}
\label{eq:gating}
\mathbf{G}
=
\tanh(\mathbf{H}_{\mathrm{pl}})
\odot
\sigma(\mathbf{X}_{\mathrm{en\_out}})
\end{equation}
Here, $\tanh(\mathbf{H}_{\mathrm{pl}})$ extracts timing-aware features where values near $\pm 1$ indicate strong peak/non-peak signals, while $\sigma(\mathbf{X}_{\mathrm{en\_out}})$ provides a soft gating mechanism with values in $(0,1)$. The element-wise product selectively amplifies encoder features at peak-relevant timesteps while preserving trend information elsewhere, acting as a conditional modulation that allows the decoder to dynamically adjust its prediction strategy according to the injected peak timing context.

A multi-head cross-attention mechanism then guides intensity decoding~\cite{vaswani2017attention}, with $\mathbf{X}_{\mathrm{de\_in}}$ as the Query and the gated context $\mathbf{G}$ as both Key and Value.
For each head $h\in\{1,\dots,H_a\}$, we compute queries, keys, and values through per-head projections:
\begin{align}
\label{eq:q}
\mathbf{Q}^{(h)} &= \mathbf{X}_{\mathrm{de\_in}}\mathbf{W}^{(h)}_{Q} \\
\label{eq:k}
\mathbf{K}^{(h)} &= \mathbf{G}\mathbf{W}^{(h)}_{K} \\
\label{eq:v}
\mathbf{V}^{(h)} &= \mathbf{G}\mathbf{W}^{(h)}_{V}
\end{align}
where \( \mathbf{W}^{(h)}_{Q}, \mathbf{W}^{(h)}_{K}, \mathbf{W}^{(h)}_{V} \in \mathbb{R}^{d \times d_h} \). Then, the aggregated attention output \( \mathbf{O}_{\mathrm{attn}} \) is obtained by concatenating all head outputs and projecting them afterward:
\begin{align}
\label{eq:attn-head}
\mathbf{O}^{(h)} &= \operatorname{softmax}\!\left( \frac{\mathbf{Q}^{(h)}(\mathbf{K}^{(h)})^{\top}}{\sqrt{d_h}} \right)\mathbf{V}^{(h)} \\
\label{eq:attn-concat}
\mathbf{O}_{\mathrm{attn}} &= \operatorname{Concat}(\mathbf{O}^{(1)},\dots,\mathbf{O}^{(H_a)})\mathbf{W}_{O}
\end{align}

Following the decoder layer, the attention output is processed through a residual connection, normalization, and a Feed-Forward Network (FFN)~\cite{vaswani2017attention}:
\begin{align}
\label{eq:ffn-1}
\mathbf{Z} &= \operatorname{LayerNorm}(\mathbf{X}_{\mathrm{de\_in}} + \mathbf{O}_{\mathrm{attn}}) \\
\label{eq:ffn-2}
\mathbf{H}_{\mathrm{dec}} &= \operatorname{LayerNorm}(\mathbf{Z} + \operatorname{FFN}(\mathbf{Z}))
\end{align}
where $\operatorname{FFN}(\cdot)$ consists of 1D convolutions with ReLU activation.
Finally, the decoder output is projected to the normalized intensity space and then inverted via RevIN, yielding the final peak-intensity forecast $\mathbf{Y}_{i\_\mathrm{pred}}$ over the forecasting horizon:
\begin{align}
\label{eq:proj}
\tilde{\mathbf{Y}}_{i\_\mathrm{pred}} &= \operatorname{Proj}(\mathbf{H}_{\mathrm{dec}}) \\
\label{eq:denorm}
\mathbf{Y}_{i\_\mathrm{pred}} &= \operatorname{DeNorm}(\tilde{\mathbf{Y}}_{i\_\mathrm{pred}}) \in\mathbb{R}^{H\times 1}
\end{align}

%%%%%%%%%%%%%%%%%%%%%%%%%%%%%%%% Unified Peak-Aware Pipeline %%%%%%%%%%%%%%%%%%%%%%%%%%%%%%%%

\subsection{Unified Peak-Aware Pipeline (UPAP)}
\label{sec:upap}

% \vspace{5pt}
\textbf{Triple Hybrid Loss Function.}
PeakFocus is trained end-to-end using a triple hybrid objective:
\begin{equation}
\label{eq:loss-total}
\mathcal{L}_{\mathrm{total}}
=
\lambda_1\mathcal{L}_{\mathrm{Global}}
+
\lambda_2\mathcal{L}_{\mathrm{Int}}
+
\lambda_3\mathcal{L}_{\mathrm{Loc}}
\end{equation}

\vspace{-2pt}

To ensure general trend consistency and guarantee basic forecasting capability, we first employ the standard Mean Squared Error (MSE) loss on the global view:
\begin{equation}
\label{eq:l-global}
\mathcal{L}_{\mathrm{Global}}
=
\operatorname{MSE}\!\left(
\mathbf{Y}_{i\_\mathrm{pred}},\;
\mathbf{Y}_{i\_\mathrm{true}}
\right)
\end{equation}

To mitigate peak sparsity and introduce tolerance awareness for the subsequent peak-specific objectives, analogous to how focal loss~\cite{lin2017focal} reweights rare classes in detection and density-based weighting addresses imbalanced regression~\cite{steininger2021density}, we construct a soft weight mask $\mathcal{W} \in [0,1]^{H\times 1}$ based on the ground-truth peak indices $\mathcal{P}=\{k\mid p_k=1\}$.
Specifically, the mask values are generated using a truncated Gaussian kernel~\cite{tobar2015learning}, as shown below:
\begin{equation}
\label{eq:y-mask}
\mathcal W
=
\max_{k\in\mathcal{P}}
\exp\!\left(-\frac{(t-k)^2}{2\gamma^2}\right)\,
\mathbb{I}(|t-k|\le\delta)
\end{equation}

Leveraging this mask, we define a peak-aware intensity objective to explicitly guide the model in capturing peak intensity. By scaling the gradients with the squared mask $\mathcal{W}^2$, this objective prioritizes high-confidence peak regions and imposes a strict focus on the peak apexes, complementing the trend supervision provided by $\mathcal{L}_{\mathrm{Global}}$:
\begin{equation}
\label{eq:l-int}
\mathcal{L}_{\mathrm{Int}}
=
\operatorname{MSE}\!\left(
\mathbf{Y}_{i\_\mathrm{pred}} \odot \mathcal{W},\;
\mathbf{Y}_{i\_\mathrm{true}} \odot \mathcal{W}
\right)
\end{equation}

Finally, we apply the same masking strategy to the temporal dimension for robust localization. Specifically, treating \( \mathcal{W} \) as the soft label for peak probability, we employ the Binary Cross-Entropy (BCE) loss to provide spatially continuous supervision and ensure effective learning, as shown below:
\begin{equation}
\label{eq:l-loc}
\mathcal{L}_{\mathrm{Loc}}
=
\operatorname{BCE}\!\left(
\mathbf{Y}_{t\_\mathrm{pred}},\;
\mathcal{W}
\right)
\end{equation}

\textbf{Tolerance-based Evaluation Protocol.}
Distinct from the soft-target training strategy, accurate assessment of peak forecasting requires mapping continuous probability outputs to discrete events. To achieve this, we establish a Tolerance-based Condense-and-Match Protocol to rigorously evaluate both localization and intensity performance.

\textbf{Condense-and-Match Mechanism.} Inspired by the tolerance-based event-matching tradition established in complex event processing~\cite{giatrakos2020cep}, our protocol operates in two stages.
To bridge continuous probabilities and discrete metrics, we first threshold $\mathbf{Y}_{t\_\mathrm{pred}}$ at $\tau$ and condense each contiguous high-probability segment to a single representative index $\hat{t}$ given by its local maximum, producing the candidate set $\hat{\mathcal{P}}$. We then enumerate all candidate pairs between $\hat{\mathcal{P}}$ and the ground-truth peak set $\mathcal{P}$ that satisfy $|t-\hat{t}| \le \delta$, sort them by temporal distance, and greedily accept unmatched pairs to enforce a strict one-to-one correspondence. Only successfully matched pairs $\mathcal{M}=\{(t, \hat{t}) \mid |t - \hat{t}| \le \delta\}$ are used for subsequent intensity error calculation. Algorithm~\ref{alg:condense-match} summarizes the full two-stage protocol used in our evaluation.

\begin{algorithm}[H]
\caption{Tolerance-based Condense-and-Match Protocol}
\label{alg:condense-match}
\small
\begin{algorithmic}[1]
\STATE \textbf{Input:} $\mathbf{Y}_{t\_\mathrm{pred}} \in [0,1]^H$, threshold $\tau$, tolerance $\delta$, and ground-truth peaks $\mathcal{P}$.
\STATE Threshold $\mathbf{Y}_{t\_\mathrm{pred}}$ at $\tau$ and split the resulting binary sequence into contiguous positive clusters.
\STATE Condense each cluster to one representative $\hat{t}$ by taking the local maximum of $\mathbf{Y}_{t\_\mathrm{pred}}$, yielding $\hat{\mathcal{P}}$.
\STATE Construct candidate pairs within the matching tolerance window $\mathcal{C}=\{(t,\hat{t},|t-\hat{t}|)\mid t\in\mathcal{P},\hat{t}\in\hat{\mathcal{P}},|t-\hat{t}|\le\delta\}$.
\STATE Sort $\mathcal{C}$ by distance and greedily accept a pair only when both $t$ and $\hat{t}$ are still unmatched.
\STATE Return $\mathcal{M}$ together with $\mathrm{TP}=|\mathcal{M}|$, $\mathrm{FP}=|\hat{\mathcal{P}}|-\mathrm{TP}$, and $\mathrm{FN}=|\mathcal{P}|-\mathrm{TP}$.
\end{algorithmic}
\end{algorithm}
\vspace{-7pt}
\textbf{Peak Timing Metrics.}
To evaluate localization precision, we report standard classification metrics derived from the matching results (i.e., unmatched items denote False Positives or False Negatives against the ground-truth peak set):
\begin{align}
\label{eq:precision}
\mathrm{Precision} &= \frac{\mathrm{TP}}{\mathrm{TP} + \mathrm{FP}} \\
\label{eq:recall}
\mathrm{Recall} &= \frac{\mathrm{TP}}{\mathrm{TP} + \mathrm{FN}} \\
\label{eq:f1}
\mathcal{F}_1 &= \frac{2 \cdot \mathrm{Precision} \cdot \mathrm{Recall}}{\mathrm{Precision} + \mathrm{Recall}}
\end{align}

\textbf{Peak Intensity Error.}
We assess regression quality exclusively on the matched set $\mathcal{M}$ (i.e., the True Positive pairs).
This design ensures that intensity errors are computed only between semantically aligned peak events, decoupling intensity estimation performance from localization failures. We define True-Positive Mean Squared Error (TP-MSE) and True-Positive Mean Absolute Error (TP-MAE) on matched peak events as:

\vspace{-10pt}

\begin{align}
\label{eq:tp-mse}
\mathrm{TP\text{-}MSE} &= \frac{1}{|\mathcal{M}|} \sum_{\mathclap{(t, \hat{t}) \in \mathcal{M}}} \left( (\mathbf{Y}_{i\_\mathrm{pred}})_{\hat{t}} - (\mathbf{Y}_{i\_\mathrm{true}})_{t} \right)^2 \\
\label{eq:tp-mae}
\mathrm{TP\text{-}MAE} &= \frac{1}{|\mathcal{M}|} \sum_{\mathclap{(t, \hat{t}) \in \mathcal{M}}} \left| (\mathbf{Y}_{i\_\mathrm{pred}})_{\hat{t}} - (\mathbf{Y}_{i\_\mathrm{true}})_{t} \right|
\end{align}

\vspace{-1pt}

\textbf{Balanced Metrics.}
To provide a comprehensive assessment, we report the Balanced Composite Score (BCS) and Peak-Intensity Magnification (PIM), which combine the unmasked TP-MSE with the localization $\mathcal{F}_1$:
\begin{flalign}
\label{eq:bcs}
\mathrm{BCS} &= \alpha(1-\mathcal{F}_1) + (1-\alpha)\left(1-\frac{1}{1+\mathrm{TP\text{-}MSE}}\right) & \\
\label{eq:pim}
\mathrm{PIM} &= \frac{1+\mathrm{TP\text{-}MSE}}{\mathcal{F}_1+\epsilon} &
% \raisetag{50pt} % 强制上提编号
\end{flalign}

\vspace{-5pt}

%%%%%%%%%%%%%%%%%%%%%%%%%%%%%%%% Experiments %%%%%%%%%%%%%%%%%%%%%%%%%%%%%%%%

\section{Experiments}
\label{sec:experiments}

%%%%%%%%%%%%%%%%%%%%%%%%%%%%%%%%%%%%%%%%%%%%%%%%%%%%%%%%%%%%%%%%
\begin{table*}[!ht]
\centering
\scriptsize
\setlength{\tabcolsep}{2.2pt}
\renewcommand{\arraystretch}{0.95}
\caption{Main results on \textsc{WLEL} and \textsc{ELC} datasets across different horizons ($H$). 
We report Recall, Precision, $\mathcal{F}_1$ for peak timing accuracy ($\uparrow$ higher is better), and TP-MSE, TP-MAE for peak intensity error ($\downarrow$ lower is better). 
BCS and PIM denote balanced metrics. We additionally integrate global metrics (MSE, Mean Absolute Error (MAE), $R^2$) into the same unified table. 
Bold indicates the best performance, and \underline{underlined} indicates the second best within each dataset-horizon block.}
\label{tab:main}

\resizebox{\textwidth}{!}{%
\begin{tabular}{l l l c c c c c c c c c c}
\toprule
\multirow{2}{*}{\textbf{Dataset}} &
\multirow{2}{*}{\textbf{$H$}} &
\multirow{2}{*}{\textbf{Model}} &
\multicolumn{3}{c}{\textbf{Peak Timing}} &
\multicolumn{2}{c}{\textbf{Peak Intensity}} &
\multicolumn{2}{c}{\textbf{Balanced Metrics}} &
\multicolumn{3}{c}{\textbf{Global Metrics}} \\
\cmidrule(lr){4-6}\cmidrule(lr){7-8}\cmidrule(lr){9-10}\cmidrule(lr){11-13}
& & &
Recall $\uparrow$ & Precision $\uparrow$ & $\mathcal{F}_1$ $\uparrow$ &
TP-MSE $\downarrow$ & TP-MAE $\downarrow$ &
BCS $\downarrow$ & PIM $\downarrow$ &
MSE $\downarrow$ & MAE $\downarrow$ & $R^2$ $\uparrow$ \\
\midrule
  
\multirow{18}{*}{\textsc{WLEL}} & \multirow{9}{*}{336}
 & CycleNet    & 0.689 & \underline{0.724} & 0.706 & 0.345 & 0.412 & 0.275 & 1.879 & 0.314 & 0.404 & 0.763 \\
& & PatchTST    & 0.616 & 0.675 & 0.644 & 0.440 & 0.485 & 0.331 & 2.202 & 0.423 & 0.485 & 0.681 \\
& & SegRNN      & 0.669 & 0.714 & 0.690 & 0.353 & 0.428 & 0.285 & 1.932 & 0.304 & 0.403 & 0.771 \\
& & STID        & 0.656 & 0.697 & 0.676 & 0.346 & 0.420 & 0.290 & 1.961 & 0.312 & 0.407 & 0.765 \\
& & TimeMixer   & 0.666 & 0.680 & 0.673 & 0.349 & 0.420 & 0.293 & 1.978 & 0.304 & 0.399 & 0.771 \\
& & Transformer & \underline{0.701} & 0.715 & \underline{0.708} & \underline{0.311} & \underline{0.394} & \underline{0.265} & \underline{1.826} & \underline{0.260} & \underline{0.370} & \underline{0.804} \\
& & Informer    & 0.633 & 0.659 & 0.646 & 0.509 & 0.516 & 0.345 & 2.303 & 0.505 & 0.533 & 0.619 \\
& & Seq2Peak    & 0.307 & 0.516 & 0.385 & 0.418 & 0.456 & 0.455 & 3.599 & 0.291 & 0.384 & 0.781 \\
& & \textbf{PeakFocus} & \textbf{0.741} & \textbf{0.770} & \textbf{0.756} & \textbf{0.264} & \textbf{0.378} & \textbf{0.227} & \textbf{1.652} & \textbf{0.221} & \textbf{0.351} & \textbf{0.833} \\
  \cmidrule(lr){2-13}
  
& \multirow{9}{*}{720}
 & CycleNet    & 0.658 & 0.674 & 0.666 & 0.579 & 0.547 & 0.351 & 2.337 & 0.514 & 0.514 & 0.601 \\
& & PatchTST    & 0.597 & 0.645 & 0.620 & 0.839 & 0.688 & 0.418 & 2.921 & 1.064 & 0.764 & 0.174 \\
& & SegRNN      & 0.648 & 0.696 & 0.671 & 0.532 & 0.545 & 0.338 & 2.253 & 0.519 & 0.537 & 0.597 \\
& & STID        & 0.647 & 0.676 & 0.661 & 0.512 & 0.530 & 0.339 & 2.253 & 0.538 & 0.542 & 0.582 \\
& & TimeMixer   & 0.671 & 0.685 & 0.678 & 0.443 & 0.482 & 0.315 & 2.101 & 0.387 & 0.456 & 0.700 \\
& & Transformer & \underline{0.714} & \underline{0.728} & \underline{0.721} & \underline{0.402} & \underline{0.459} & \underline{0.283} & \underline{1.919} & \underline{0.347} & \underline{0.430} & \underline{0.731} \\
& & Informer    & 0.629 & 0.663 & 0.646 & 0.702 & 0.630 & 0.383 & 2.596 & 0.664 & 0.627 & 0.485 \\
& & Seq2Peak    & 0.256 & 0.438 & 0.324 & 0.504 & 0.542 & 0.505 & 4.517 & 0.372 & 0.463 & 0.711 \\
& & \textbf{PeakFocus} & \textbf{0.748} & \textbf{0.770} & \textbf{0.759} & \textbf{0.317} & \textbf{0.415} & \textbf{0.240} & \textbf{1.712} & \textbf{0.262} & \textbf{0.382} & \textbf{0.797} \\
  \midrule
  
  \multirow{18}{*}{\textsc{ELC}} & \multirow{9}{*}{336}
 & CycleNet    & 0.419 & 0.418 & 0.418 & \underline{0.699} & 0.653 & 0.497 & 3.974 & \underline{0.358} & 0.445 & \underline{0.650} \\
 & & PatchTST    & \underline{0.710} & \underline{0.766} & \underline{0.737} & 1.111 & 0.805 & 0.394 & 2.826 & 0.458 & 0.491 & 0.552 \\
& & SegRNN      & 0.606 & 0.651 & 0.628 & 0.811 & 0.691 & 0.410 & 2.845 & 0.393 & 0.465 & 0.616 \\
 & & STID        & 0.366 & 0.387 & 0.377 & \textbf{0.579} & \textbf{0.602} & 0.495 & 4.112 & 0.359 & 0.453 & \underline{0.650} \\
 & & TimeMixer   & 0.693 & 0.729 & 0.711 & 0.850 & 0.705 & \underline{0.374} & \textbf{2.568} & \textbf{0.344} & \textbf{0.430} & \textbf{0.665} \\
& & Transformer & 0.382 & 0.314 & 0.345 & 0.726 & 0.680 & 0.537 & 5.266 & 0.469 & 0.516 & 0.543 \\
& & Informer    & 0.541 & 0.423 & 0.475 & 0.703 & 0.639 & 0.469 & 3.518 & 0.518 & 0.550 & 0.495 \\
 & & Seq2Peak    & 0.470 & 0.500 & 0.485 & 0.709 & \underline{0.636} & 0.465 & 3.456 & 0.361 & \underline{0.441} & 0.647 \\
 & & \textbf{PeakFocus} & \textbf{0.719} & \textbf{0.777} & \textbf{0.747} & 0.969 & 0.758 & \textbf{0.372} & \underline{2.600} & 0.443 & 0.482 & 0.567 \\
  \cmidrule(lr){2-13}

 & \multirow{9}{*}{720}
 & CycleNet    & 0.435 & 0.427 & 0.431 & \underline{0.726} & 0.668 & 0.495 & 3.915 & \underline{0.388} & \underline{0.476} & \underline{0.619} \\
 & & PatchTST    & 0.706 & 0.748 & \underline{0.726} & 1.068 & 0.806 & 0.393 & 2.795 & 0.531 & 0.546 & 0.479 \\
& & SegRNN      & \underline{0.708} & 0.741 & 0.724 & 0.793 & 0.686 & \textbf{0.346} & \textbf{2.358} & 0.409 & 0.489 & 0.598 \\
 & & STID        & 0.457 & 0.480 & 0.468 & \textbf{0.652} & \underline{0.638} & 0.463 & 3.467 & \textbf{0.387} & 0.481 & \textbf{0.620} \\
 & & TimeMixer   & 0.674 & 0.678 & 0.676 & 0.840 & 0.713 & \underline{0.390} & \underline{2.694} & 0.410 & 0.483 & 0.598 \\
& & Transformer & 0.530 & 0.420 & 0.469 & 0.929 & 0.755 & 0.505 & 4.194 & 0.611 & 0.572 & 0.400 \\
 & & Informer    & 0.334 & 0.244 & 0.282 & 0.727 & \textbf{0.607} & 0.549 & 5.322 & 0.992 & 0.778 & 0.026 \\
 & & Seq2Peak    & 0.500 & 0.532 & 0.516 & 0.794 & 0.671 & 0.463 & 3.412 & 0.403 & \textbf{0.468} & 0.604 \\
 & & \textbf{PeakFocus} & \textbf{0.712} & \textbf{0.751} & \textbf{0.731} & 1.221 & 0.870 & 0.408 & 2.996 & 0.641 & 0.554 & 0.370 \\
\bottomrule
\end{tabular}}
\vspace{-10pt}
\end{table*}
%%%%%%%%%%%%%%%%%%%%%%%%%%%%%%%%%%%%%%%%%%%%%%%%%%%%%%%%%%%%%%%%

%%%%%%%%%%%%%%%%%%%%%%%%%%%%%%%% Experimental Setup %%%%%%%%%%%%%%%%%%%%%%%%%%%%%%%%

\subsection{Experimental Setup}
\label{sec:setup}

\textbf{Datasets.} We evaluate PeakFocus on two complementary real-world hourly load datasets, summarized in Table~\ref{tab:datasets}. Both datasets are univariate load series; ground-truth peaks are obtained offline using the detector defined in Section~\ref{sec:peak_def} with the lookahead $\ell$ listed in Table~\ref{tab:hyperparameters}.

(1) \textbf{\textsc{ELC}}~\cite{zhou2021informer}: A widely-used public benchmark adopted in the standard preprocessed Electricity Load Diagrams version released by Informer, containing 26{,}304 consecutive hourly records, with 1{,}429 structural peaks (5.43\% of timestamps). The series captures aggregated consumer-level consumption with irregular usage patterns and stronger high-frequency local fluctuations. Following the chronological split convention adopted by prior TSF benchmarks~\cite{zhou2021informer, nie2023time}, the series is partitioned into contiguous training, validation, and test windows along the temporal axis without overlap or shuffling.

(2) \textbf{\textsc{WLEL}}: World Large-scale Electricity Load is an industrial-scale hourly electricity load dataset collected from a real regional grid, spanning 2021-01-01 to 2025-09-25 (41{,}476 consecutive hourly records, 2{,}944 structural peaks, 7.10\% of timestamps). Compared with consumer-level benchmarks, it exhibits clear daily/weekly periodicity and seasonal modulation typical of grid-aggregated load. The training window covers 2021-01 to 2023-08. The validation window (2023-09--2024-08) spans a complete annual cycle for stable model selection, and the test window (2024-09--2025-09) covers a full subsequent year for long-term forward-deployment robustness assessment. This dataset is constructed under a research collaboration between our team and a large-scale industrial partner in the power sector, providing genuine grid-aggregated load measurements from an in-operation industrial deployment; under a confidentiality agreement, the raw load series cannot be publicly released, but all preprocessing configurations and peak-detection hyperparameters (Table~\ref{tab:hyperparameters}) are fully disclosed to ensure reproducibility of the experimental protocol on equivalent industrial deployments.\looseness=-1

\begin{table}[t]
\centering
\small
\caption{Statistics of the two evaluation datasets.}
\label{tab:datasets}
\setlength{\tabcolsep}{4pt}
\renewcommand{\arraystretch}{1.05}
\begin{tabular*}{\columnwidth}{@{\extracolsep{\fill}}lcc}
\toprule
\textbf{Property} & \textbf{ELC} & \textbf{WLEL (ours)} \\
\midrule
Source          & Informer-released ECL  & Real regional grid \\
Granularity     & Hourly                 & Hourly \\
Time span       & 2016-07 to 2019-07     & 2021-01 to 2025-09 \\
Samples      & 26{,}304               & 41{,}476 \\
Peak ratio      & 5.43\% (1{,}429)       & 7.10\% (2{,}944) \\
\bottomrule
\end{tabular*}
\vspace{-22pt}
\end{table}

\textbf{Baselines.} We evaluate PeakFocus against two categories of state-of-the-art (SOTA) models: TSF models and ELPF models.
The TSF baselines include Transformer~\cite{vaswani2017attention}, Informer~\cite{zhou2021informer}, PatchTST~\cite{nie2023time}, SegRNN~\cite{lin2025segrnn}, CycleNet~\cite{lin2024cyclenet}, STID~\cite{shao2022spatial}, and TimeMixer~\cite{wang2024timemixer}. To ensure a fair comparison, these TSF models are adapted to our UPAP framework, incorporating dual-head outputs and the triple hybrid loss (Eq.~\ref{eq:loss-total}).
For the ELPF baseline, we select Seq2Peak~\cite{zhang2023unlocking}, which augments the Transformer backbone by incorporating a parameter-free max-pooling decoder to extract peak-hour series and employs a hybrid loss function to jointly optimize for both global reconstruction and peak intensity.
Furthermore, we validate the universal effectiveness of our optimization strategy by comparing the vanilla versions of these TSF models against their UPAP-enhanced counterparts, which we report in the Generality Analysis (Section~\ref{sec:ablation}) of the ablation study.

\begin{itemize}\setlength{\itemsep}{1pt}
\item Transformer~\cite{vaswani2017attention} uses vanilla self-attention to map historical tokens to future tokens.
\item Informer~\cite{zhou2021informer} uses ProbSparse sparse attention to reduce quadratic cost and decode the full horizon in one pass.
\item PatchTST~\cite{nie2023time} uses non-overlapping patches with channel-independent processing to extract local patterns.
\item SegRNN~\cite{lin2025segrnn} uses a segment-level recurrent unit to shorten recurrence depth on long horizons.
\item CycleNet~\cite{lin2024cyclenet} uses a learnable cycle component and forecasts only the residual.
\item 
STID~\cite{shao2022spatial} uses multi-scale decomposition with linear mixing of trend and seasonal parts.
\item 
TimeMixer~\cite{wang2024timemixer} uses multi-scale decomposition with linear mixing of trend and seasonal parts.
\item 
Seq2Peak~\cite{zhang2023unlocking} uses a Transformer backbone with a max-pooling decoder to extract peak-hour series and optimize global reconstruction and peak intensity.
\end{itemize}

%%%%%%%%%%%%%%%%%%%%%%%%%%%%%%%% Implementation Details %%%%%%%%%%%%%%%%%%%%%%%%%%%%%%%%
% \vspace{-12pt}
\subsection{Implementation Details}
\label{sec:implementation}

We implement PeakFocus using PyTorch and conduct all experiments on a single NVIDIA A100 GPU.
To ensure statistical reliability, all experiments are repeated 5 times with different random seeds, and we report the mean and standard deviation.
The model is optimized using the Adam optimizer with an initial learning rate of $10^{-3}$.
The training process runs for a maximum of 20 epochs with a batch size of 128, employing early stopping with a patience of 5 epochs.
For the triple hybrid objective in Eq.~\ref{eq:loss-total}, we set the loss weights to $\lambda_1=0.2$ for global MSE, $\lambda_2=0.4$ for TP-MSE, and $\lambda_3=0.4$ for localization BCE. PeakFocus uses a latent dimension $d_{\text{model}}=256$, a 2-layer residual MLP backbone, a dual-scale MSM-PL pyramid ($K=2$), pooling configurations $(\kappa^{(1)},s^{(1)})=(3,2)$ and $(\kappa^{(2)},s^{(2)})=(5,2)$, and 1D convolutional feature extractors with kernel size $\kappa_c=3$ and GELU activation. The LAD uses a feed-forward dimension $d_{\text{ff}}=256$ with ReLU activation. Table~\ref{tab:hyperparameters} summarizes the main hyperparameter settings used in all experiments.

\begin{table*}[t]
    \caption{Main-text hyperparameter settings for PeakFocus.}
    \label{tab:hyperparameters}
    \centering
    \small
    \renewcommand{\arraystretch}{1.1}
    \begin{tabular*}{\textwidth}{@{\extracolsep{\fill}}llll}
    \toprule
    \multicolumn{2}{l}{\textbf{Training Dynamics \& Loss Function}} & \multicolumn{2}{l}{\textbf{Model Architecture \& Inference}} \\
    \cmidrule(r){1-2} \cmidrule(l){3-4}
    Input Sequence Length ($L$)              & 168          & Model Dimension ($d_{\text{model}}$)                     & 256 \\
    Prediction Horizon ($H$)                 & \{336, 720\} & Feed-Forward Dim ($d_{\text{ff}}$)                       & 256 \\
    Batch Size                               & 128          & Attention Heads ($n_{\text{heads}}$)                     & 4 \\
    Learning Rate                            & 0.001        & Encoder Layers ($e_{\text{layers}}$)                     & 1 \\
    Optimizer                                & Adam         & MLP Layers (Backbone)                                    & 2 \\
    Training Epochs                          & 20           & Dropout                                                  & 0.1 \\
    LR Scheduler                             & Warm-up 3 + ExpDecay $0.9^{e-3}$ & Activation (LAD / MSM-PL)                  & ReLU / GELU \\
    Model Saving Criterion                   & Lowest BCS   & MSM-PL Pooling $k$=1 ($\kappa^{(1)}/s^{(1)}$)       & 3 / 2 \\
    Global Loss Weight ($\lambda_1$)         & 0.2          & MSM-PL Pooling $k$=2 ($\kappa^{(2)}/s^{(2)}$)       & 5 / 2 \\
    TP-MSE Loss Weight ($\lambda_2$)         & 0.4          & MSM-PL Conv Kernel $\kappa_c$                            & 3 \\
    Localization Loss Weight ($\lambda_3$)   & 0.4          & Peak Lookahead ($\ell$)                                  & 3 (WLEL) / 5 (ELC) \\
    Gaussian Kernel Width ($\gamma$)         & 1.0          & Peak Tolerance ($\delta$)                                & 1 \\
    BCS Weight ($\alpha$)                    & 0.5          & Probability Threshold ($\tau$)                           & 0.4 \\
    PIM Constant ($\epsilon$)                & 0.01         & Sensitivity Threshold ($\eta$)                           & 0 (\texttt{findpeaks} default) \\
    \bottomrule
    \end{tabular*}
    \vspace{-10pt}
\end{table*}

%%%%%%%%%%%%%%%%%%%%%%%%%%%%%%%% Main Results Analysis %%%%%%%%%%%%%%%%%%%%%%%%%%%%%%%%
\subsection{Main Results Analysis}
\label{sec:main_results}
Table~\ref{tab:main} presents the unified comparison across all baselines and both forecasting horizons, reporting peak-event quality, intensity error, and global-reconstruction accuracy under a single evaluation protocol for direct cross-method comparison.

\textbf{Peak Forecasting Quality.}
On WLEL, PeakFocus consistently achieves the best peak timing ($\mathcal{F}_1$) and intensity estimation (TP-MSE, BCS) at both horizons, outperforming the strongest baseline by approximately 7\% in $\mathcal{F}_1$ and 15\% in TP-MSE. This validates the core design: UPAP, MSM-PL, and LAD together address the three bottlenecks identified in Section~\ref{sec:intro}. Seq2Peak, which adopts a fixed max-pooling surrogate for peak localization, lags significantly behind, confirming that structurally defined peaks require learnable localization rather than heuristic aggregation.

\textbf{Cross-Dataset Generalization.}
PeakFocus is positioned as an ELPF-specific framework rather than a general-purpose TSF model, so we prioritize peak-event metrics ($\mathcal{F}_1$, TP-MSE, BCS, PIM) over global reconstruction. The two datasets here represent two complementary regimes: WLEL captures grid-aggregated load with strong daily and seasonal periodicity, where peaks emerge as well-shaped structural events; ELC captures consumer-level demand with substantially more high-frequency noise and irregular individual usage patterns, where peaks are less regular and more easily masked by local fluctuations. Across both datasets and both horizons ($H$=336 and $H$=720), PeakFocus achieves the best $\mathcal{F}_1$, Recall, and Precision; in contrast, competitive baselines exhibit clearly larger $\mathcal{F}_1$ fluctuations once the regime shifts from the structured WLEL to the noisier ELC at the same horizon. This suggests that the peak-aware design generalizes across regimes rather than being tied to dataset-specific patterns.

\textbf{Global Reconstruction and the ELC Trade-off.}
On WLEL, the triple hybrid objective improves rather than harms global metrics: PeakFocus attains the best MSE, MAE, and $R^2$ in addition to all peak metrics, leading on every reported indicator. On ELC, dedicated global-fit models such as TimeMixer and STID achieve lower MSE and MAE than PeakFocus. Our work targets a complementary problem, predicting when and how high the next peak occurs, which directly informs reserve scheduling and overload-risk mitigation; the corresponding peak-event metrics are the primary indicators in our setting, and PeakFocus consistently leads on them across both datasets. In the deployment-relevant grid-aggregated regime of WLEL, peak-aware training and global reconstruction reinforce each other; on the noisier consumer-level ELC, the two objectives are harder to optimize jointly, which further motivates ELPF as a separate task complementary to general-purpose TSF.

\begin{figure}[t]
  \centering
  \includegraphics[width=\columnwidth]{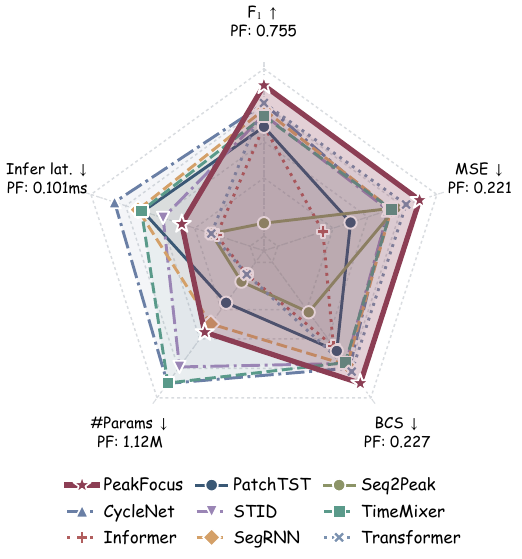}
  \vspace{-15pt}
  \caption{Efficiency radar on WLEL ($H$=336). Five axes compare quality ($\mathcal{F}_1$, MSE, BCS) and efficiency (\#Params, inference latency). PeakFocus leads all quality axes while remaining competitive on both parameter count and inference latency.}
  \label{fig:radar}
  \vspace{-15pt}
\end{figure}

\textbf{Efficiency Analysis.}
As shown in Figure~\ref{fig:radar}, we provide a holistic quality--efficiency comparison on WLEL ($H$=336) across five axes. PeakFocus leads all three quality axes ($\mathcal{F}_1$, MSE, BCS) with only 1.12M parameters, while lightweight models such as CycleNet (114K) and TimeMixer (116K) are more parameter-efficient but substantially weaker in peak-event quality. This suggests that the gain of PeakFocus does not come from simply increasing model capacity, but from allocating a still-compact model budget to peak-aware localization and regression modules that better match the sparse-event nature of ELPF to meet an accuracy-efficiency balance.

%%%%%%%%%%%%%%%%%%%%%%%%%%%%%%%% Ablation Study %%%%%%%%%%%%%%%%%%%%%%%%%%%%%%%%
\subsection{Ablation Study}
\label{sec:ablation}

\begin{figure*}[!htbp]
\centering
\includegraphics[width=0.9\textwidth]{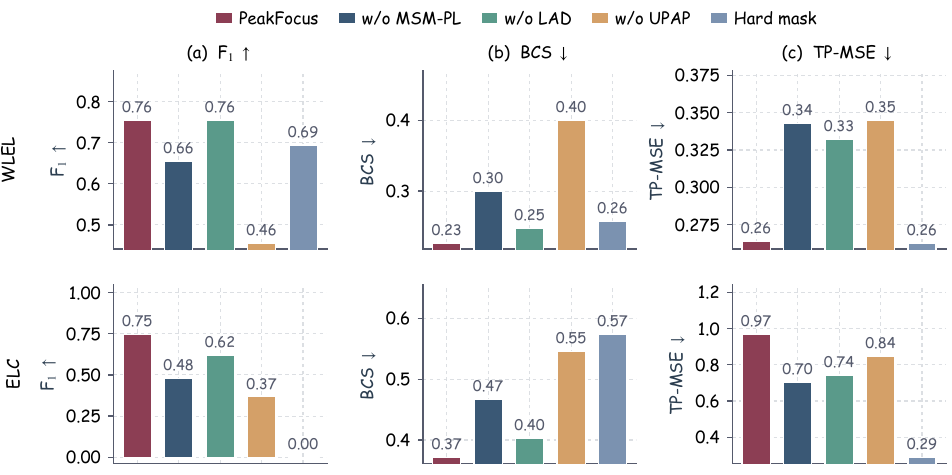}
\vspace{-5pt}
\caption{Ablation study on WLEL and ELC ($H$=336). Five variants: PeakFocus (full); w/o MSM-PL (linear layer replacing pyramid); w/o LAD (no peak timing injection); w/o UPAP (standard MSE only); Hard mask (binary mask replacing soft Gaussian). Metrics: $\mathcal{F}_1\uparrow$, BCS$\downarrow$, TP-MSE$\downarrow$; values averaged over 5 seeds.}
\label{fig:ablation_combined}
\vspace{-10pt}
\end{figure*}

\begin{figure*}[!htbp]
\centering
\includegraphics[width=0.9\textwidth]{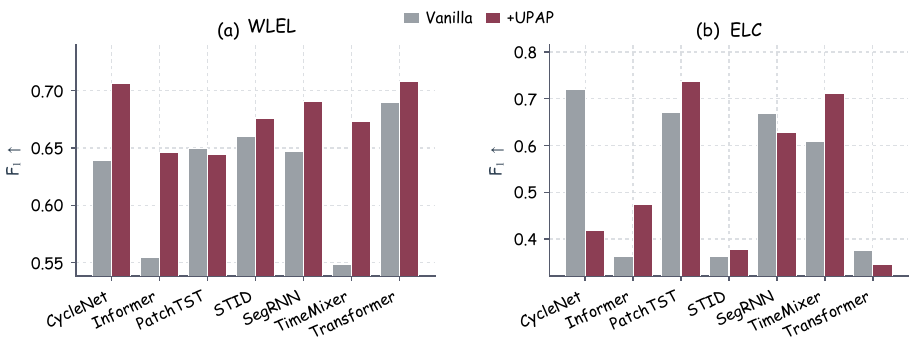}
\vspace{-5pt}
\caption{Cross-backbone generality of UPAP on WLEL and ELC ($H$=336). Each grouped bar compares the vanilla baseline (Vanilla, gray) against the same backbone enhanced with +UPAP (red) on $\mathcal{F}_1\uparrow$. On WLEL, +UPAP consistently improves all seven baselines. On ELC, most backbones benefit, though CycleNet and Transformer show slight degradation on more volatile consumer-level data, revealing a dataset-dependent trade-off on volatile consumer-level data that is worth further investigation.}
\label{fig:upap_generality}
\vspace{-6pt}
\end{figure*}

\begin{figure*}[!htbp]
\centering
\includegraphics[width=\textwidth]{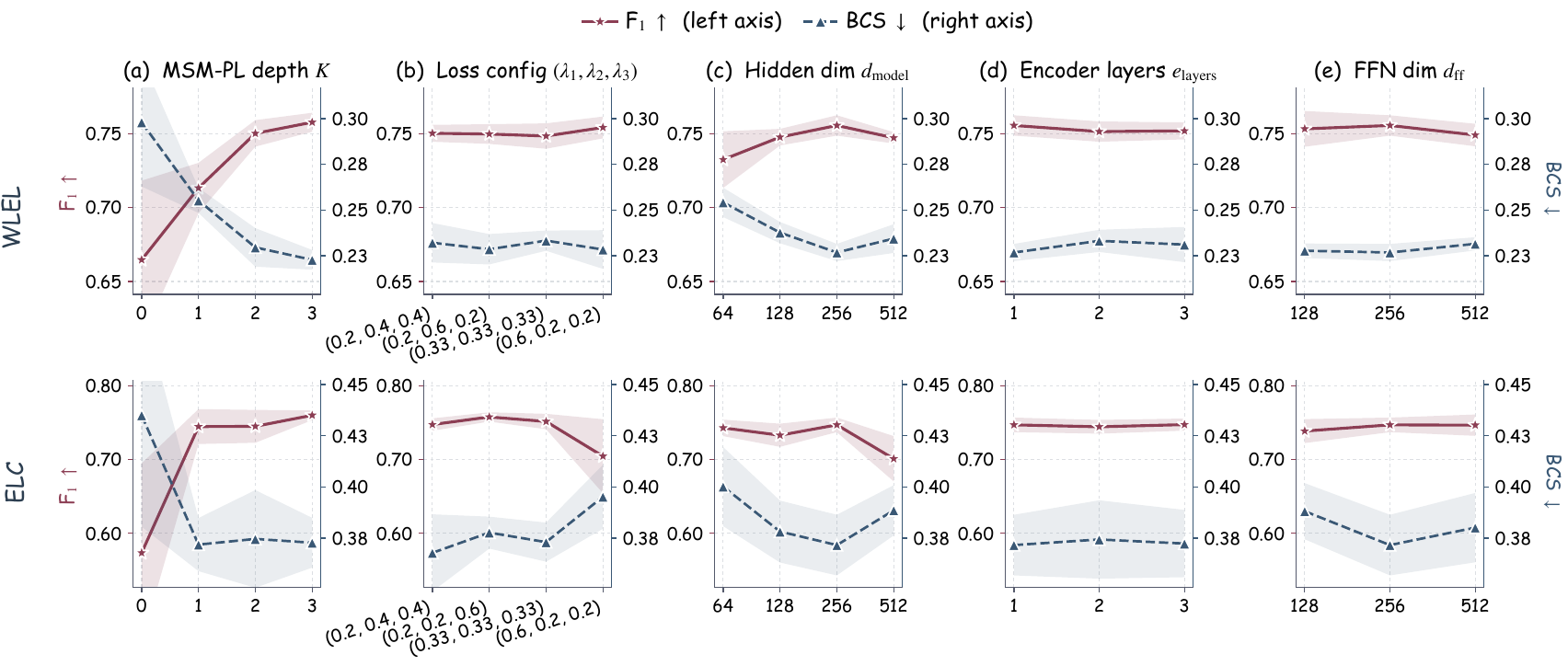}
\vspace{-15pt}
\caption{Parameter sensitivity on WLEL (top) and ELC (bottom). (a)~MSM-PL depth $K$. (b)~Loss configuration $(\lambda_1,\lambda_2,\lambda_3)$: C1=(0.2,0.4,0.4), C2=(0.2,0.6,0.2), C3=(0.33,0.33,0.33), C4=(0.6,0.2,0.2). (c)~Hidden dimension $d_{\text{model}}$. (d)~Encoder layers $e_{\text{layers}}$. (e)~FFN dimension $d_{\text{ff}}$. All curves are averaged over five runs with different random seeds for robustness assessment.}
\label{fig:param_sensitivity}
\vspace{-10pt}
\end{figure*}

We perform component analysis to validate each proposed module. Figure~\ref{fig:ablation_combined} summarizes the ablation results on both datasets, Figure~\ref{fig:upap_generality} examines the cross-backbone generality of UPAP, and Figure~\ref{fig:param_sensitivity} examines parameter sensitivity.

\textbf{Observation 1: Multi-Scale Cascade Is Essential for Peak Localization.}
We validate the core design of MSM-PL from two complementary perspectives. First, as shown in Figure~\ref{fig:ablation_combined}, completely replacing MSM-PL with a simple linear layer (w/o MSM-PL) causes $\mathcal{F}_1$ to drop by $\sim$13\% on WLEL with a pronounced decline in Precision, confirming that single-scale processing cannot suppress local fluctuations. Second, as shown in Figure~\ref{fig:param_sensitivity}(a), varying the cascade depth $K$ reveals a progressive improvement: $K$=0 (no cascade) yields the lowest $\mathcal{F}_1$ on both datasets, as the model lacks the coarse-grained semantics needed to filter local fluctuations; increasing $K$ to 2 enables the top-down cascade to inject multi-scale features and substantially improves both timing precision and balanced accuracy; $K$=3 yields only marginal additional gains. Together, these results corroborate the multi-scale representation conflict identified in Section~\ref{sec:intro}: the hierarchical coarse-to-fine cascade is indispensable for mitigating misjudgment and resolving temporal misalignment, and a dual-scale pyramid ($K$=2) achieves the best accuracy--capacity trade-off consistently.\looseness=-1

% Table ablation removed — replaced by Figure ablation_combined with enriched caption.

\textbf{Observation 2: Explicit Timing Context Improves Intensity.}
As shown in Figure~\ref{fig:ablation_combined}, removing the peak hidden state injection (w/o LAD) preserves localization $\mathcal{F}_1$ but notably increases TP-MSE on both datasets. This validates the intensity smoothing hypothesis from Section~\ref{sec:intro}: without explicit peak timing context, the decoder lacks awareness of where peaks occur and consequently underestimates their intensity under the pull of global smoothing trends from the MSE objective.

\textbf{Observation 3: Unified Optimization Bridges Localization and Regression.}
As shown in Figure~\ref{fig:ablation_combined}, removing the triple hybrid objective (w/o UPAP) causes the most severe degradation among all variants, with $\mathcal{F}_1$ dropping by over 40\% on WLEL and BCS nearly doubling. This directly addresses the paradigm disconnect motivation: standard MSE optimization is dominated by non-peak data, causing the model to neglect sparse extreme events. UPAP counteracts this bias through synchronized timing--intensity supervision.

UPAP is also model-agnostic. As shown in Figure~\ref{fig:upap_generality}, applying UPAP consistently improves $\mathcal{F}_1$ across all seven backbones on WLEL, with the largest gains on weaker vanilla models. On the more volatile ELC, most backbones still benefit, though models with strong inherent periodicity modeling (e.g., CycleNet) show degradation, since CycleNet’s explicit cycle encoding already captures periodic patterns well, and the additional peak-timing gradient may disturb this stable periodic modeling rather than provide extra useful information.\looseness=-1

\textbf{Observation 4: Soft Gaussian Masks Provide Stable Supervision.}
Replacing the soft Gaussian mask with a hard binary mask degrades $\mathcal{F}_1$ and BCS on WLEL, and causes complete collapse on ELC ($\mathcal{F}_1 \approx 0$). The soft mask provides tolerance-aware gradient smoothing around peak positions, which is critical for robust learning when peak timing annotations contain inherent uncertainty, particularly on more volatile consumer-level data. This underscores the importance of tolerance-aware supervision for sparse peaks.

% Table mask_type removed — Hard mask rows merged into Table ablation (red rows).

\textbf{Observation 5: Robustness to Architectural Hyperparameters.}
We examine four structural hyperparameters to assess the stability of PeakFocus under different internal configurations. As shown in Figure~\ref{fig:param_sensitivity}(b)--(e): (i)~all four loss weight configurations yield comparable $\mathcal{F}_1$ and BCS, indicating that the triple hybrid objective is insensitive to the specific weighting scheme; (ii)~$d_{\text{model}}$ improves from 64 to the default 256 and then plateaus, with $d_{\text{model}}$=512 even degrading $\mathcal{F}_1$ on ELC by $\sim$6\%, suggesting that PeakFocus does not rely on large model capacity; (iii)~varying $e_{\text{layers}}$ from 1 to 3 or $d_{\text{ff}}$ across \{128, 256, 512\} produces negligible changes on both datasets. These results demonstrate that PeakFocus is robust to moderate architectural variations, and the default configuration ($d_{\text{model}}$=256, $e_{\text{layers}}$=1, $d_{\text{ff}}$=256) is a stable and effective choice.

%%%%%%%%%%%%%%%%%%%%%%%%%%%%%%%%%%%%%%%%%%%%%%%%%%%%%%%%%%%%%%%%

%%%%%%%%%%%%%%%%%%%%%%%%%%%%%%%%%%%%%%%%%%%%%%%%%%%%%%%%%%%%%%%%

%%%%%%%%%%%%%%%%%%%%%%%%%%%%%%%% Qualitative Visualization %%%%%%%%%%%%%%%%%%%%%%%%%%%%%%%%
\subsection{Qualitative Visualization}
\label{sec:qualitative_visualization}

\begin{figure*}[!t]
\centering
\includegraphics[width=\textwidth]{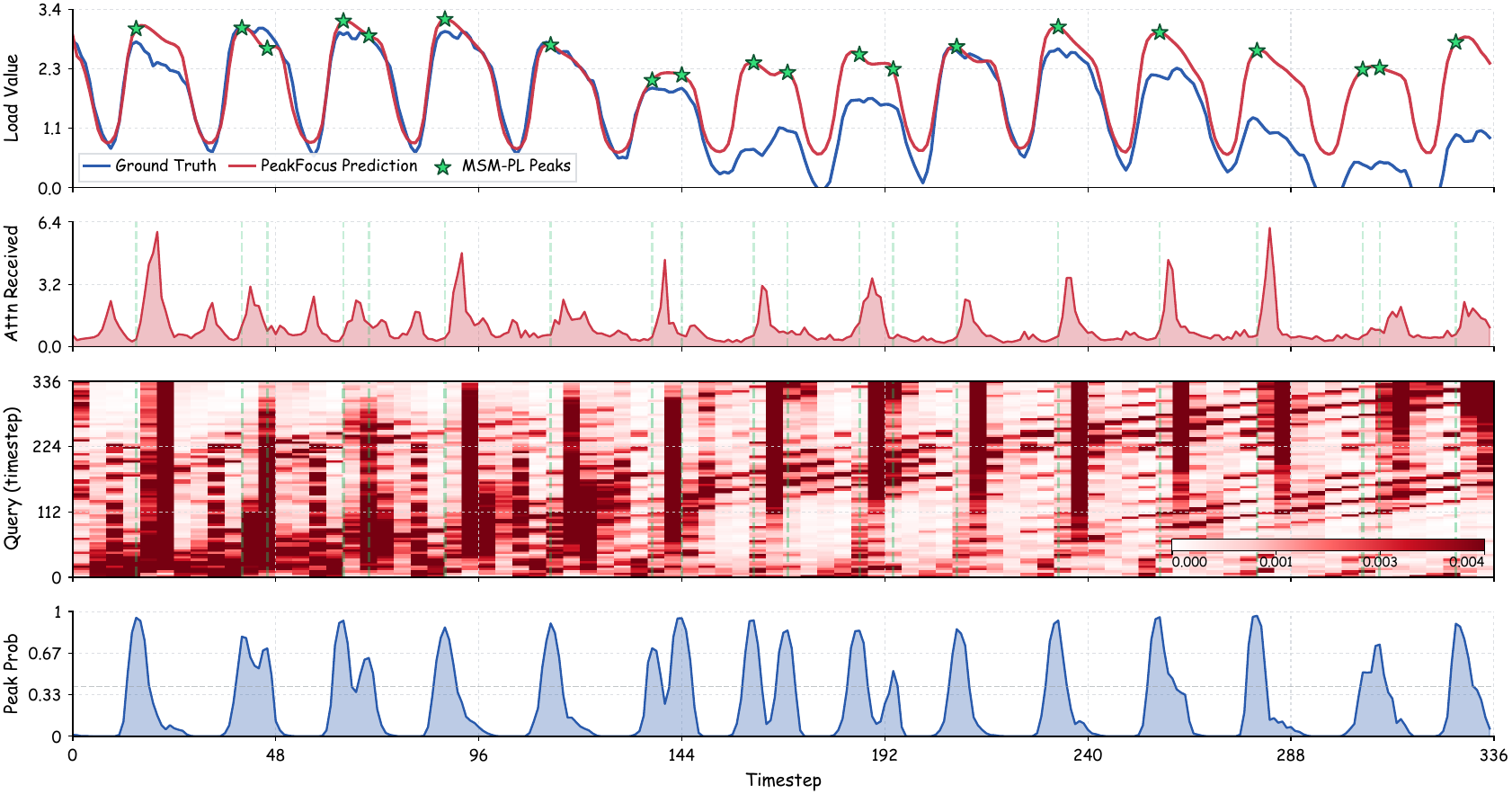}
\vspace{-15pt}
\caption{Internal mechanism visualization on WLEL ($H$=336). From top to bottom: (1)~Ground truth vs.\ PeakFocus prediction with MSM-PL peaks (green stars); (2)~Aggregated LAD cross-attention per timestep, concentrated at peaks; (3)~Attention heatmap (query $\times$ key) with bright vertical bands of peak-relevant attention; (4)~MSM-PL peak probability with detection threshold (dashed). The alignment of peak probability, attention, and intensity prediction confirms the localization-to-regression pipeline inside the proposed framework and improves interpretability for peak-aware load forecasting decisions.}
\label{fig:gate_vis}
\vspace{-6pt}
\end{figure*}

\begin{figure*}[!t]
  \centering
  \begin{subfigure}{0.49\textwidth}
    \centering
    \includegraphics[width=\linewidth]{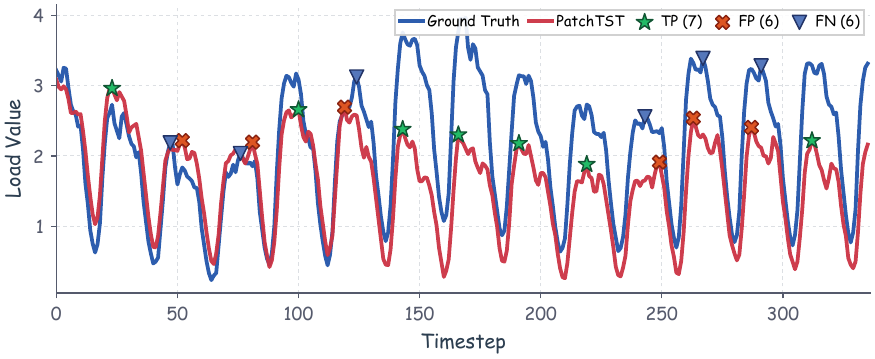}
    \caption{PatchTST: Timing Misalignment + Intensity Smoothing}
    \label{fig:patchtst_7500}
  \end{subfigure}
  \hfill
  \begin{subfigure}{0.49\textwidth}
    \centering
    \includegraphics[width=\linewidth]{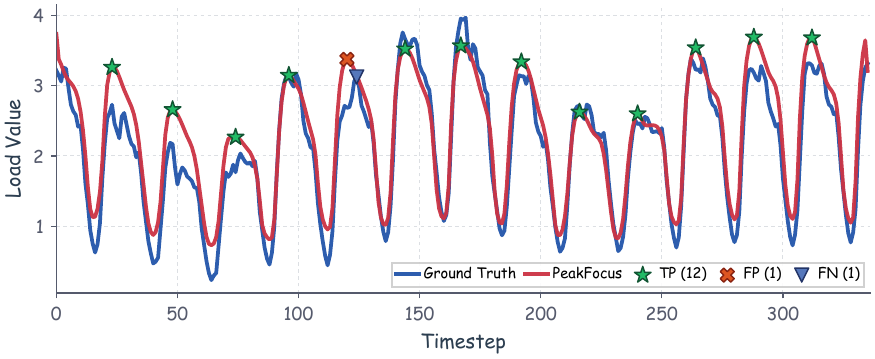}
    \caption{PeakFocus: Precise Localization + Faithful Intensity}
    \label{fig:peakfocus_7500}
  \end{subfigure}

  \vspace{-5pt}
  \caption{Qualitative comparison on the WLEL dataset ($H$=336). (a)~PatchTST suffers from both timing misalignment (TP=7, FP=6, FN=6) and systematic intensity smoothing, where peak intensities are underestimated by an average of 0.80 due to the global MSE objective diluting sparse peak signals. (b)~PeakFocus achieves near-perfect localization (TP=12, FP=1, FN=1) and faithfully tracks peak intensities with lower TP-MSE, demonstrating that the joint localization--regression design simultaneously resolves both bottlenecks. Green stars: True Positives; red crosses: False Positives; blue triangles: False Negatives.}
  \label{fig:vis_comparison}
  \vspace{-10pt}
\end{figure*}

We conduct a qualitative comparison between PeakFocus and PatchTST on a representative WLEL test sample to visually verify how the proposed modules jointly address the localization and intensity bottlenecks identified in Section~\ref{sec:intro}.

\textbf{Localization and Intensity Analysis (Figure~\ref{fig:vis_comparison}).}
As shown in Figure~\ref{fig:vis_comparison}a, PatchTST exhibits both timing misalignment and intensity smoothing simultaneously. On the localization side, it achieves only 7 True Positives against 6 False Positives and 6 False Negatives, indicating that the model frequently misjudges non-peak fluctuations as peaks while missing genuine structural peaks. On the intensity side, PatchTST systematically underestimates peak intensities by an average of 0.80, producing a flattened trajectory that fails to capture extreme load surges, which is a direct consequence of the global MSE objective diluting sparse peak signals toward the mean trend. In contrast, Figure~\ref{fig:vis_comparison}b demonstrates that PeakFocus simultaneously resolves both bottlenecks. The MSM-PL achieves near-perfect localization (TP=12, FP=1, FN=1) by leveraging the hierarchical pyramid to suppress local fluctuations and the cascade injection to preserve temporal alignment. Meanwhile, the LAD faithfully tracks peak intensities with lower TP-MSE, because it explicitly conditions regression on the peak timing context, counteracting intensity smoothing. This joint improvement validates that coupling localization and regression in a unified framework enables each component to reinforce the other.

\textbf{Analysis of Internal Attention Mechanism (Figure~\ref{fig:gate_vis}).}
To further verify how LAD leverages peak timing context, we visualize the internal states of a representative WLEL sample ($H$=336) in Figure~\ref{fig:gate_vis}. The top panel shows the ground truth and PeakFocus prediction with MSM-PL detected peaks (green stars). The second panel displays the aggregated attention received by each timestep in the LAD cross-attention layer: attention clearly concentrates at peak positions, confirming that the decoder selectively focuses on peak-relevant moments. The attention heatmap (third panel) further reveals that queries across all timesteps predominantly attend to key positions near detected peaks, forming distinct vertical activation bands. The bottom panel shows the peak probability output from MSM-PL, which exhibits sharp spikes well-aligned with true peak locations. This cross-panel consistency also indicates that the attention weights are not diffuse correlations alone, but are anchored by explicit locator outputs that guide regression toward operationally critical peak hours. This visualization demonstrates that MSM-PL provides reliable timing signals and LAD effectively translates them into targeted intensity estimation, validating the design rationale of our two-stage localization-to-regression pipeline.

%%%%%%%%%%%%%%%%%%%%%%%%%%%%%%%% Related Works %%%%%%%%%%%%%%%%%%%%%%%%%%%%%%%%
\section{Related Works}
\label{sec:related}

\subsection{Time Series Forecasting (TSF)}
TSF plays a pivotal role in various domains, evolving from Recurrent Neural Networks (RNNs) to MLPs and Transformers~\cite{yu2025large, zhu2025breaking}.
RNN-based approaches, SegRNN~\cite{lin2025segrnn} utilizes recurrent units for efficient long-term modeling, while P-sLSTM~\cite{kong2025unlocking} improves upon Long Short-Term Memory (LSTM) by incorporating patching and channel independence.
Meanwhile, MLP-based models, including DLinear~\cite{zeng2023transformers}, STID~\cite{shao2022spatial}, N-BEATS~\cite{oreshkin2020nbeats}, and TimeMixer~\cite{wang2024timemixer}, achieve strong performance via simple spatial-temporal decomposition.
Transformers like Informer~\cite{zhou2021informer} and Autoformer~\cite{wu2021autoformer} reduce complexity using sparse attention, FEDformer~\cite{zhou2022fedformer} leverages frequency-domain decomposition, whereas PatchTST~\cite{nie2023time} and iTransformer~\cite{liu2024itransformer} improve results through patch-based tokenization and inverted attention. TimesNet~\cite{wu2023timesnet} further captures intra-period and inter-period variations by reshaping 1D series into 2D tensors, Pyraformer~\cite{liu2022pyraformer} introduces pyramidal attention for multi-resolution temporal patterns, and SCINet~\cite{liu2022scinet} extracts hierarchical temporal dependencies through recursive sample convolutions.
These backbones aggregate short-range and long-range context for the dominant trend, but no module preserves an explicit signal of where extreme values occur, so local maxima are still collapsed during decoding.
Within the data-engineering community, MAGNN~\cite{chen2023magnn} exploits a multi-scale pyramid graph network to preserve inter-variable dependencies across time scales, BasicTS+~\cite{shao2024basicts} delivers a unified benchmarking infrastructure that exposes dataset heterogeneity as a first-class dimension, and TimeKD~\cite{liu2025timekd} performs calibrated language-model distillation for efficient multivariate forecasting, together reflecting the rising interest in deployable time series data management.
Recent work targets deployment-oriented forecasting~\cite{cirstea2022towards, miao2024unified, zhao2024multistep} and rare-deviation isolation in time series~\cite{kieu2022robust}, objectives close in spirit to peak detection.
However, these general TSF models are often suboptimal for the rigorous demands of ELPF. Since peaks constitute sparse tail events in the load distribution, standard objectives drive models to learn a conservative representation biased towards the mean to minimize aggregate penalty. Consequently, critical sharp spikes are effectively treated as negligible fluctuations and smoothed out~\cite{hutimefilter}.
Even multi-scale models like TimeMixer~\cite{wang2024timemixer} and AMD~\cite{hu2025adaptive} prioritize global trend decomposition, lacking the fine-grained resolution required for precise peak timing. These limitations motivate task-specific designs that supervise peak timing and intensity jointly to learn the pattern.

\subsection{Electricity load peak forecasting (ELPF)}
Methodologies have shifted from heuristic two-stage predict-then-locate paradigms to unified end-to-end frameworks~\cite{amara2023daily, dai2021electrical}, where conventional pipelines treat detection as post-processing and optimize features for average error rather than peak discrimination.
Recent end-to-end approaches bridge this gap: Seq2Peak~\cite{zhang2023unlocking} uses a max-pooling decoder, and Seq2LPP~\cite{zhu2025enhancing} adds weather-guided attention.
Hu~et~al.~\cite{hu2025metaload} transfers peak-load patterns across regional grids via meta-learning, but still optimizes a global regression loss rather than supervising peak events directly.
However, defining peaks as fixed-interval maxima reduces ELPF to downsampling and fails to capture structural peaks from local topology. Existing methods also struggle with the multi-scale representation conflict and insufficiently condition intensity regression on peak timing context, leading to intensity smoothing.
These limitations indicate that existing ELPF studies have moved toward end-to-end modeling in form, but not yet in event semantics: peaks remain weakly represented, loosely aligned, and insufficiently injected into downstream intensity estimation.
PeakFocus addresses this gap by treating peak timing as a learning signal and propagating it into the intensity decoder.

\section{Conclusion}
\label{sec:conclusion}

PeakFocus bridges temporal localization and intensity regression in ELPF via a unified end-to-end framework: MSM-PL resolves peak misjudgment and timing misalignment through hierarchical multi-scale fusion, and LAD prevents intensity smoothing by conditioning regression on explicit peak timing context. By coupling timing supervision and intensity estimation in a single optimization, the framework addresses peak misjudgment, timing misalignment, and intensity smoothing in concert rather than in isolation. Experiments on the public ELC benchmark and the industrial-scale WLEL dataset confirm a new state of the art across timing, intensity, and balanced metrics. Beyond the proposed model, the Tolerance-based Condense-and-Match Protocol offers a reusable evaluation methodology for sparse structural-event forecasting. Future work includes capacity scaling on longer industrial deployments, probabilistic peak forecasting, and integration with live grid telemetry for real-time scheduling.

\FloatBarrier
\section*{AI-Generated Content Acknowledgment}
AI-assisted language tools (e.g., ChatGPT) were employed solely for grammar correction and wording adjustment. All the research ideas, experimental design, and reported results were independently produced and verified by authors.
% \clearpage
\bibliographystyle{IEEEtran}
\bibliography{references}

\end{document}